\pdfoutput=1

\documentclass{article}
\usepackage[utf8]{inputenc}
\usepackage{fullpage}
\usepackage{url}
\usepackage{xcolor}
\usepackage[colorlinks = true,
            linkcolor = blue,
            urlcolor  = blue,
            citecolor = blue,
            anchorcolor = blue]{hyperref}
\usepackage{amsmath}
\usepackage{enumitem}
\newcommand{\subscript}[2]{$#1 _ #2$}
\usepackage{graphicx}
\usepackage{multirow}

\usepackage{listings}
\usepackage{fancyvrb}
\usepackage{bera}

\usepackage{fancyhdr}
\usepackage[margin=1in,headsep=.2in]{geometry}
\fancyhfoffset[L]{0.5cm} 
\fancyhfoffset[R]{0.5cm} 
\title{COVID-19: Strategies for Allocation of Test Kits}
\author{Arpita Biswas\thanks{arpitab@iisc.ac.in, Indian Institute of Science}\and Shruthi Bannur\thanks{bshruthi.137@gmail.com} \and Prateek Jain\thanks{prajain@microsoft.com, Microsoft Research India}\and Srujana Merugu\thanks{srujana.merugu@gmail.com}}
\date{April 2, 2020}

\begin{document}

\maketitle

\begin{abstract}
With the increasing spread of COVID-19, it is important to systematically test more and more people. The current strategy for test-kit allocation is mostly rule-based~\cite{icmr2020}, focusing on individuals having (a)~symptoms for COVID-19, (b)~travel history or (c)~contact history with confirmed COVID-19 patients. Such testing strategy may miss out on detecting asymptomatic individuals who got infected via community spread. Thus, it is important to allocate a separate budget of test-kits per day targeted towards preventing community spread and detecting new cases early on. 

In this report, we consider the problem of allocating test-kits and discuss some solution approaches. We believe that these approaches will be useful to contain community spread and detect new cases early on. Additionally, these approaches would help in collecting unbiased data which can then be used to improve the accuracy of machine learning models trained to predict COVID-19 infections.\\


\noindent{\bf DISCLAIMER:} This report suggests
 some easy-to-implement test allocation strategies based on discussions among individual researchers. We have not empirically tested out these strategies in the current context and do not make any claims regarding the efficiency or safety of strategies. All the suggestions are individual thoughts of authors and the associated institutions do not endorse them or recommend them. 
\end{abstract}

\section{Background}
\begin{figure}[h]
\centering
        \includegraphics[scale=0.5]{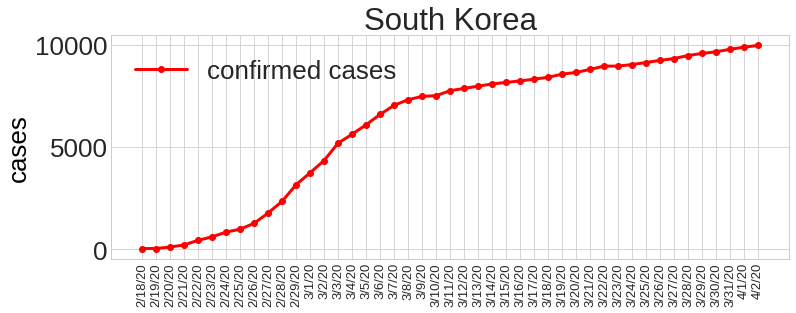}
    \caption{The cumulative number of confirmed cases and deaths in South Korea. We clearly see the flattening of confirmed cases in South Korea. Data source~\cite{data-source}}
    \label{fig:trend-SK}
\end{figure}

\begin{figure}[h]
\centering
        \includegraphics[scale=0.5]{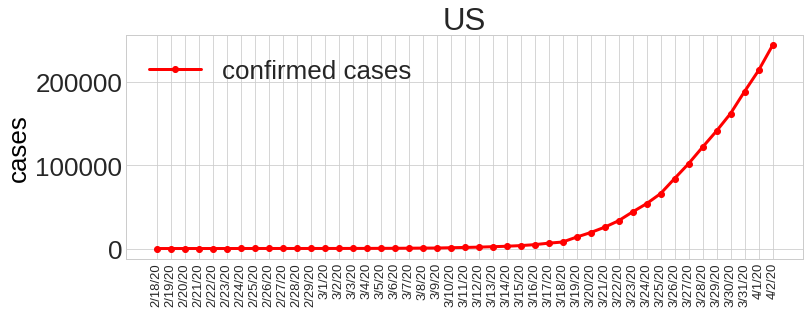}
    \caption{The cumulative number of confirmed cases and deaths in United States (US). We observe an exponential increase of confirmed cases in US. Data source~\cite{data-source}}
    \label{fig:trend-US}
\end{figure}

South Korea, a country of 50 million people, has set an example of successfully flattening the curve of new COVID-19 infections by conducting over 400,000 tests~\cite{nyt2020flatten} (Figure~\ref{fig:trend-SK}). This was achieved by setting up drive-through testing, allowing at least 10,000 people to be tested per day. South Korea’s foreign minister Kang Kyung-wha, in an interview with BBC News~\cite{bbc2020coronavirus}, said that “Testing is central because that leads to early detection, minimizes further spread, and quickly treats those found with the virus”. Several countries are suffering from severe community spread because of their delays in testing~\cite{nyt2020delay}, two of the prime examples being the United States and Italy. In the United States, among a population of 330 million, the number of confirmed cases is more than 230,000 with over 10,000 deaths and these numbers are growing exponentially~(Figure~\ref{fig:trend-US}), whereas in South Korea there are around 9976 confirmed cases and 169 deaths (as of April 2, 2020). Thus, early testing and repeated testing at regular intervals are two of the key strategies to ensure a low fatality rate. 
However, for countries with a large population (more than 100 million), it is difficult to adopt exhaustive testing schemes because of the limited number of available testing-kits and facilities. Testing a lot of people with mild or no symptoms would occupy the limited testing resources, which could otherwise be used for high-risk patients. However, it is also important to test individuals with mild or no symptoms to detect asymptomatic cases~\cite{livemint2020icmr} and implement a method that systematically tests individuals for COVID-19. Such systematic testing methods can be designed with various intents, namely,
(a)~early detecting among health workers and other essential-service workers, 
(b)~containment as well as prevention of community spread, (c)~timely care for the ones who are at potential risk, and (d)~collecting data for improving the risk of infection assessment models and learning counterfactual scenarios, which is essential to determine medical demand, economic impact and policy making.

\section{Current Testing Policy and Gaps}
As per the new guidelines by the Indian Council of Medical Research (ICMR)~\cite{icmr2020}, the current testing strategy mandates testing of all symptomatic contacts of confirmed cases, all symptomatic health workers, and all patients with fever, cough, and shortness of breath. For asymptomatic individuals with recent travel history, ICMR provides strict guidelines to be home-quarantined and get tested when symptomatic. All the asymptomatic direct contacts of confirmed cases are asked to get tested once between 5 and 14 days. Although the current testing method is targeted towards containing the COVID-19 infection, there are some gaps:
\begin{enumerate}
    \item The known cases are mostly self-reported. Thus, the testing happens only among the symptomatic cases and misses asymptomatic cases that do not report their contact with confirmed cases.
    \item Asymptomatic patients who get infected via community spread are missed.
    \item The data collected using the current testing strategy is heavily biased towards patients with symptoms of infections. This makes the data inappropriate to be used directly for training an infection-prediction model. 
\end{enumerate}
In particular, the asymptomatic cases are the most important factors for community spread. Since these cases are hard to detect, the current testing policies may not be adequate to contain the spread of the virus~\cite{nyt2020asymptomatic}. \textbf{Thus, it is important to allocate a separate budget, $\mathbf{K}$ test-kits per day, targeted towards (a)~detecting new cases early on and also (b)~collecting unbiased data}. We formally define the \textit{test allocation problem} in Section~\ref{sec:problem}. Next, in Section~\ref{sec:proposal}, we propose some strategies for allocating a fixed number of test-kits across the population each day. 

\section{Test Allocation Problem}\label{sec:problem}
We assume that, on each day $i$, a budget of $K_i$ is declared for containing community spread, where $K_i$ represents the number of test-kits available on the next day, i.e., $(i+1)^{th}$ day. We consider the problem of selecting a set of $K_i$ individuals on that day, and recommend them to take a test (for detecting whether he/she is infected). 

\subsection{Problem Statement}
We assume that there is a symptom-checker app, where individuals can fill in their personal information (such as age, gender), along with medical history (diabetes, cardiovascular diseases), and travel information (country traveled within the last 14 days). The individuals are assumed to use the symptom-checker app at any time of the day, and the symptom checker is required to decide whether to recommend testing to an individual. Note that this data could be filled by the individuals on their own or by field-level health workers (ASHAs~\cite{asha}) or even obtained from legacy health survey datasets. The symptom-checker coordinates\footnote{The coordination between symptom checker app and local testing centers would facilitate the symptom checker to allocate a dedicated time-slot to each individual for taking the test. This would prevent large gatherings at the test centers. In this paper, we do not tackle the problem of allocating fixed time-slots to individuals. However, this remains an interesting direction for future research.} with the local testing facilities to estimate the number of test-kits available (for additional recommendations) on the next day. There can be two modes of recommendation:
\begin{enumerate}
    \item \textit{Offline mode}: Recommendation can be delayed till the end of the day, after all the users have registered the symptom checker on a particular day $i$. 
    \item \textit{Online mode}: Recommendation is made as soon as the individual enters his/her information in the app.
\end{enumerate}
The selected individuals are then supposed to undergo a test\footnote{Ensuring the individuals take the test with the right incentives, mandates, and ethical safeguards is yet another direction for further study.}. It would typically take one more day to observe the test-result (infected or not). 

\noindent\textbf{Goal}: Given a budget of $K_i$ test-kits, design a strategy to select at most $K_i$ individuals each day such that the observations (infected or not) for those individuals help in improving the predictive power of an infection-prediction model. This goal is aligned with the need to observe enough number of individuals having distinct feature vectors. Thus, such a strategy is implicitly targeted towards collecting unbiased data as well as detecting new cases early on (even among asymptomatic cases) which in turn would help in containing the community spread.



\subsection{Notation}

Let $X^{all}$ represent the entire population of the country (irrespective of whether they have filled-in their details in the app). Let $X$ denote the set of all individuals who fill in their details in the symptom-checker app, possibly on different days.  Clearly, $X^{all}$ is a superset of $X$. Let each individual $x\in X$ be denoted by a vector of $d$ features $x=(x_1,x_2,\ldots,x_d)$. These features capture demographic and health-related information for each individual. If an individual $x$ take the test, then the label $y(x)$ denotes the test-result. $y(x)=1$ denotes that $x$ is infected and $y(x)=0$ denotes that $x$ is not infected. Thus, $X$ can be represented as a matrix with rows representing individuals and columns representing features as well as the label $y$. 

Let $X_i$ be the set of individuals who fill in their details in the symptom checker app on the day $i$. Let $S_i\subset X_i$ be the set of $K_i$ individuals who are selected. We assume that $K_i<<X_i$ (for most practical scenarios). Note that the $y(x)$ labels for all $x\in S_i$ will eventually be available. Let $S$ be the set of all the individuals who took the test. Note that for all $x\in S$ the corresponding $y$ labels\footnote{In certain cases, an individual might be asked to take a test on multiple days for accuracy. That is still counted as a single test and the associated label is assumed to be positive if any of the tests are positive.} are observed.
On the other hand, the labels $y(x)$ will be unavailable for all $x\notin S$, which is the case for most individuals in  $X$ and $X^{all}$. 

In this work, we focus on the \textit{offline mode} and propose some approaches in Section~\ref{sec:proposal}. We briefly discuss the online mode and other important considerations in Section~\ref{sec:discussion}.

\section{Solution Approaches: Offline Mode}\label{sec:proposal}
In the offline mode, the symptom checker is allowed to delay the recommendation of testing until the end of the day. This mode of operation would require the users to register before a specific time of the day (say, $9 PM$) to be eligible for testing on the next day. Let the set of all eligible users on the day $i$ be denoted as $X_i$. Within $10-15$ minutes of this deadline (i.e., by $9:15$ PM), the selected $K_i$ individuals (denoted as $S_i \subseteq X_i$) would then be notified about getting the test done on the next day, i.e., $i+1^{th}$ day.

In this section, we propose four approaches, namely~(a)~randomization across four disjoint buckets, (b)~stratification based on features, (c)~budget delayed contextual bandits, and (d)~utility-based active learning solution. These solutions are aimed at selecting $K_i$ individuals among the population $X_i$ who fills in their details in the symptom checker on a day $i$. The first two approaches ($a$ and $b$) are easy to implement and can be quickly adopted for allocating test-kits, while the next two approaches ($c$ and $d$) are more sophisticated and yield better results in long term. 

\subsection{Randomization across four disjoint buckets}
A naive approach is to assign each individual in population $X_i$ to one of the following four mutually disjoint groups. 
\begin{enumerate}[label=\subscript{X}{{(\arabic*)}}:]
    \item symptomatic cases with risky  contact or travel history (i.e., contact with confirmed cases or international travel), 
    \item asymptomatic cases with risky contact or travel history,
    \item symptomatic cases without any risky travel or contact history, and
    \item asymptomatic cases without any risky travel or contact history.
\end{enumerate}
Split the $K_i$ test-kits among four buckets: allocate a budget of $K_{X_{(j)}}$ to each group $j\in \{1,2,3,4\}$ such that $\sum_j K_{X_{(j)}} = K_i$. The value $K_{X_{(j)}}$ depends on how much importance should be given to the group $X_{(j)}$ while allocating testing-kits. Then, randomly sample from each group $j$ based on the budget allocation for that group. 

Note that testing should remain mandatory for all those in the critical group (symptomatic cases with either recent travel history or contact history). Since there is an overlap with the mandatory cases, the budget $K_i$ can be adjusted accordingly.

\subsection{Stratification based on features}\label{sec:stratification}
In this approach, we wish to select both the under-represented and over-represented groups in the population $X_i$ to the same extent as that in the larger population $X^{all}$, especially when the features are not distributed evenly across the set $X_i$ (as observed by symptom-checker).

Let us assume that each person in the population has a feature vector $x$ which includes age as one of the features. A naive random test allocation strategy is the one in which we sample uniformly the population who have filled out the symptom checker (i.e., $X_i$). In this allocation strategy, the distribution across the sampled features would mimic the distribution across $X_i$. For example, if $90\%$ of the rows of $X_i$ contain individuals less than 60 years old, then the random allocation strategy is likely to select individuals who are less than $60$ years old and might miss out on the individuals above $60$ years. We wish to avoid this scenario and ensure that there is adequate representation for the $60+$ population in the chosen sample.

To sample across all features according to the distribution of features in $X^{all}$, we adopt the stratification approach. In this approach, we divide the population space (both $X$ and $X^{all}$) into  groups (i.e., disjoint subsets) based on the values of some \textit{selected} features (represented as $F$). If we know the joint distribution of the population over the set of features in $X^{all}$ and $X_i$, we can use them to adjust the weight of each individual according to the group he/she falls in. This means that, if the overall population $X^{all}$ is uniformly distributed across all groups and an individual $x\in X$ falls in a group with fewer samples in $X_i$, the weight, or probability of selecting that individual, will be higher than for group with many samples of $X_i$.

Formally, we define the overall population as $X^{all}$, where each individual is represented by a vector $x = [x_{1},x_{2}, \ldots, x_{d}]$. Let $F\subset\{1,\ldots,d\}$ be a selected set of features based on which the population is grouped. The joint distribution of the features in $F$ over the population $X^{all}$ be denoted as $P_{X^{all}}(x_f = a_f;\ \forall f\in F)$ where $a_f$ is the value that a feature $f \in F$ can take. Similarly, let $P_{X_i}(x_f = a_f;\ \forall f\in F)$ be the joint distribution for the population $X_i$ (set of individuals who have reached out to the symptom check on day $i$). We can use these values to assign a weight~\cite{selection-bias} on an individual $x=[x_1,\ldots,x_d]$ which is proportional to the value $\frac{P_{X^{all}}(x_f = a_f;\ \forall f\in F)}{P_{X_i}(x_f = a_f;\ \forall f\in F)}$.  

Moreover, we also increase the weights of individuals who are at a higher risk. To find the utility values, let us assume that there is an external function $U(x)$ that outputs the utility associated with testing an individual $x$; see Section~\ref{sec:utility} for more discussions on $U(\cdot)$. 

The weight on each individual $x$ can be computed as 
\[weight(x) = U(x)*\frac{P_{X^{all}}(x_f = a_f;\ \forall  f\in F)}{P_{X_i}(x_f = a_f;\ \forall f\in F)}\]

This weight would then be used to perform a weighted random sampling of $K_i$ individuals from the pool of $X_i$ individuals. 

\textbf{Example}: let us consider two features, namely, age and gender, for each individual in the population\footnote{We can do a similar stratification across geographical regions, e.g., states/districts, past medical conditions, and  occupations}. We can denote the joint distribution of age and gender over the entire population by $P_{X^{all}}(Age, Gender)$, which can be obtained using the information provided in the National Health Profile (see Table~\ref{StateAgeGender} for state-wise distribution of age and gender or Table~\ref{AgeGender} for the overall distribution of age and gender) and the Census. We wish to select samples from $X_i$ such that the selected set of samples is representative of the population $X^{all}$ (either for a given state or for the overall country). 

Now, we divide the population $X_i$ into groups, for example $P_{X_i}(Age<20, Gender:Female)$, $P_{X_i}(Age<20, Gender:Male)$, $P_{X_i}(20<Age<40, Gender:Female)$, $P_{X_i}(20<Age<40, Gender:Male)$, etc. These values represent the density of the population that falls within that bucket. When selecting among the population, we will weight the samples with the inverse of these values. Similarly, we would use  $P_{X_{all}}(\cdot)$

Putting all these together, we provide a simple pseudo-code below.

\begin{Verbatim}[commandchars=\\\{\}]
# Compute the fraction of X_i who belong to a particular (Gender, Age) category pair 
> for each gender in [Male, Female]:
    for each age in [<20, 20-40, 40-60, 60-80, >80]
        joint_users[gender, age] = count_of_samples_in(X_i, gender, age)/number_of_rows(X_i)
        
# Let the joint probabilities be estimated for Gender and Age categories in X^{all}
# using information provided in National Health Profile (using Table \ref{StateAgeGender} or \ref{AgeGender}):
> for each gender in [Male, Female]:
    for each age in [<20, 20-40, 40-60, 60-80, >80]:
        joint_overall[gender, age]=proportion(X^{all}, gender, age)
        
# Create a 1-D list of unique IDs       
> ids = extract_unique_ids(X_i) 

# Assign weights to each individual depending on their Gender and Age categories
> weighted_probabilities = list(size=length(ids)) # Initialize a list of weights
> for j in 1 to length(ids): 
    gender = X_i[j, "Gender"]
    age = X_i[j, "Age"]
    risk = U(X[j,]) # Assume function U(x) outputs risk score for each individual x
    weighted_probabilities[j] = risk * joint_overall[gender, age]/joint_users[gender, age]
    
# Choose K_i items from the list of ids using weighted random sampling
> select = choice(list_of_candidates = ids,
                number_of_items_to_pick = K_i,
                adjusted_probability = weighted_probabilities,
                sample_with_replacement = False)
\end{Verbatim}
Notes for implementation: 
\begin{itemize}
    \item the joint probabilities, \texttt{joint\_users[gender, age]} may have zero value for certain (gender, age) pairs and computation of corresponding users $j$ (\texttt{weighted}\_\texttt{probabilities[j]}) would throw an error. This be adjusted by assigning $0$ to \texttt{weighted}\_\texttt{probabilities[j]} whenever \texttt{joint\_users[gender, age]} equals $0$, or use back-off methods for smoothing~\cite{smoothing}. 
    \item the Python package \texttt{numpy.random.choice} generates a weighted random sample of a fixed size from a given array.
\end{itemize}




\subsection{Budgeted delayed contextual bandits:}
In this approach, we combine the classifier learning with “informative” data point selection. The algorithm’s goal is to iteratively refine the classifier so that the cost of missed detections or false positives is similar to the best you could have done in hindsight. 

In a contextual bandits setting, the decision-maker observes a context, makes a decision by choosing one action from a number of alternative actions and later observes a reward associated with that decision. The goal is to maximize the average reward.
In our case, the context is information about the user: their age, geolocation, prior health conditions, symptoms, travel history, contact history, etc. The actions are: to test a person or not. The reward or cost is: a) cost of missing detection (i.e. not testing a person when she/he is actually infected) and b) cost of false alarm (i.e. testing a person when she/he is not infected). 

In general, this algorithm constructs a confidence ball around every point and its prediction (i.e. where the person should be tested or not). The final decision per point is sampled from a probability distribution over both the outcomes: positive, negative. Here is a tutorial and an implementation of this algorithm:
https://vowpalwabbit.org/tutorials/contextual\_bandits.html

Above mentioned, Vowpal Wabbit~\cite{vowpal} is an open-source library that implements online and offline training algorithms for contextual bandits. Offline training and evaluation algorithms are described in the paper titled “Doubly Robust Policy Evaluation and Learning”~\cite{dudik2011doubly} by Dudik et al.

For implementation, we can bootstrap the algorithm using the stratification approach on the dataset obtained until day $i-1$. After this, we model the problem as a contextual bandits problem day $i$ onward. 

\begin{figure}[h]
\centering
        \includegraphics[scale=0.7]{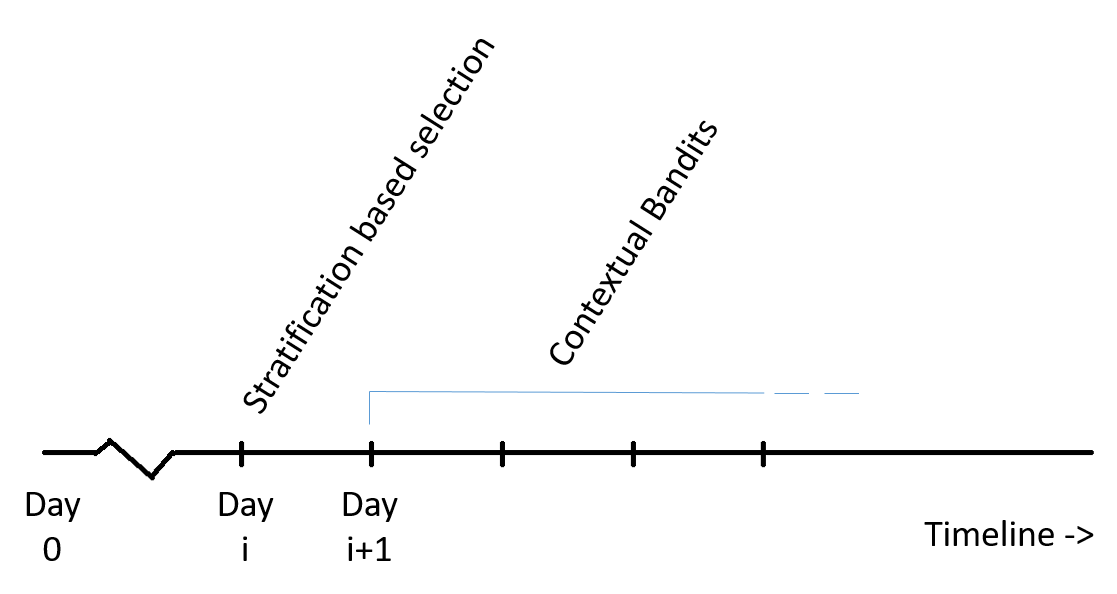}
    \caption{Illustrating budgeted delayed contextual bandits approach}
    \label{fig:contextual}
\end{figure}

The main components of the problem are:
\begin{itemize}
    \item \textbf{Context}: information about an individual, such as, age, gender, geolocation, prior health conditions, present symptoms, travel history, and contact history, represented as a vector $x$ with $d$ dimensions. 
    \item \textbf{Actions}: to test a person (arm $1$) or not (arm $0$). This is a 2-arms bandit setting ($A=\{0,1\}$).
    \item \textbf{Feedback}: Whether a person is positive ($Y=\{0,1\}$).
    \item \textbf{Reward/Cost}: reward/cost of testing a person when he/she is positive $C(x;a=1,y=1)$ and when he/she is not positive $C(x; a=1, y=0)$. We can fix cost/reward of not testing to be $0$ (i.e. $C(x; a=0, y)=0$ for both $y=1,0$.
    \item \textbf{Policy}: takes a context $x$ and returns an action $a$ with probability $\mathcal{P}(a|x)$. We denote by $P_i()$, the policy used on day $i$.
    \item \textbf{Budget}: the number of available test-kits, say $K_i$.
    \item \textbf{Decision to optimize}: learn a policy that selects $K_i$ individuals to maximize the expected reward
\end{itemize}

\noindent Now, let $X_i$ represent the set of new data points collected on the day $i$ for which we need to decide the subset $S_i$ that should go for testing. Recall that size of $S_i$ should be bounded by the budget $K_i$. That is, $|S_i|\leq K_i$. 

Also, let $\widetilde{Z}_{i-1}$ be the set of points for which we had asked for testing and some section of them are tested and their results are available (we assume that the feedback $y$, whether or not the selected individuals are positive, is delayed by a day). That is,  $\widetilde{Z}_{i-1}=\{(x_e, y_e), \forall e\}$ where $y_e$ is the test result of individual with feature vector (context) $x_e$.  The algorithm works as follows:
\begin{enumerate}
    \item \textbf{Update}: We use $\widetilde{Z}_{i-1}$ and the cost function to train (or re-train) a classifier and update scoring function $f()$ using standard contextual bandit update methods.
    $f(x)$ gives a score for testing individual $x$; higher score implies one of the two scenarios (a)~the individual is more likely to be positive or (b)~the algorithm is confused and is trying to build confidence around $x$. 
    \item \textbf{Prediction}:  We apply updated function $f$ on each individual $x\in X_i$ to get their scores $f(x), x\in X_i$. These risk probabilities and the associated costs act as input to a contextual bandit algorithm~\cite{vowpal} and it outputs the probability of recommending testing, $P_i(a=1|x)$. 
    \item \textbf{Selection}: The next task is to select a total of $K_i$ individuals on day $i$. We select using a weighted random sampling with recommendation probabilities $P_i(a=1|x)$ (as described in the pseudocode above).
\end{enumerate}

\subsection{Utility-based active learning}
In this approach, the goal is to find a set of most “informative” data points (people) so that getting them tested (i.e. labeling them) would lead to a significant increase in risk assessment solution performance, i.e. the classifier that predicts if a person is likely to be positive and hence should be tested.

There are two types of solutions to this problem: 
\begin{itemize}
    \item \textbf{Uncertainty based exploration}: Determine which user’s infection status is most uncertain according to the current classifier’s scoring function and ask for their labels (see, for example, the paper by Tong and Koller, 2001~\cite{tong2001support}), but this method will suffer in our case with high bias in the initial classifier.
    \item \textbf{Disagreement based exploration}: In this approach, if any two classifiers in the hypothesis class disagree on whether a person is likely to be infected or not, then that person should be tested. However, this algorithm is somewhat challenging to implement. One version of the algorithm is implemented here~\cite{krishnamurthy2019active}. 
\end{itemize}

\section{Discussions}\label{sec:discussion}
\subsection{Solution Approach: Online Mode}
The challenge in the \textit{online mode} is that individuals arrive online one-by-one, and one needs to take an immediate decision on whether to recommend them for testing. To do that, we need to design a rate-limiting mechanism~\cite{rate-limiting-google-cloud}. At a high level, a rate-limiter limits the number of actions that can be
performed within a particular time window. To apply a rate-limiter mechanism, let us divide a day into six time-slots of $4$ hours each and limit recommending only $\alpha_t$ fraction of $K_i$ at a particular time-slot $t$. The value of $\alpha_t$ can be estimated by computing the proportion of individuals who registered at a particular slot on the previous day, i.e.,
\[\alpha_t = \frac{\#count\_of\_arrivals(slot=t, \ day=(i-1))}{|X_{i-1}|}.\] 

For an individual $x$ who arrives during the time-slot $t$ on the day $i$, we compute the score (probability or utility of recommendation) using one of the four approaches defined in the offline mode. If there were less than $(\alpha_t*K_i)$ individuals on the previous day, then recommend testing to $x$. If not, then compare $x$'s score with that of the $(\alpha_t*K_i)^{th}$ highest scorer of the previous day among those who arrived in slot $t$. If $x$ scores higher than the $(\alpha_t*K_i)^{th}$ scorer on the previous day, then recommend testing. 

Apart from rate-limiting mechanisms, there exist other sophisticated techniques that can be applied to facilitate the online mode of operation.

\subsection{Coordination Between Symptom Checker and Testing Facilities}
The test-allocation problem that we consider assumes that the symptom checker app coordinates with local testing centers to estimate the number of available test-kits on the next day. Such coordination is also important for allowing the symptom checker to allocate a dedicated time-slot to each individual for taking the test. This would prevent large gatherings at the test centers. In this paper, we do not tackle the problem of allocating fixed time-slots to individuals. However, this remains an interesting direction for future research.

\subsection{Finding the Associated Utility}\label{sec:utility}
In Section~\ref{sec:stratification}, we assume an external function $U(x)$ that outputs the utility associated with an individual $x$.  This utility  function has to be chosen based on the health policy directives~\cite{icmr2020}.  Note that this utility value can be  computed using all the available information (e.g., location, symptoms, occupation) associated with an individual, including features that might not have been used for stratification. When the budget $K_i$ itself forms a big fraction of the overall test kits, it is important to address the care and containment aspects along with data collection requirements.  Let $M$ denote the current risk assessment model. Then $U(x)$ set to the  probability of infection  $p_M(y=1|x)$  or a monotonic function thereof is a good choice since it  ensures that the high risk cases are assigned more test-kits while allowing a non-zero allocation for the low risk ones. When the high risk patients are already guaranteed tests and the goal for the $K_i$ budget is to improve the infection assessment model for a binary decision, then it would be appropriate to choose $U(x)$  to be the entropy of the conditional distribution $H(p_M(y|x))$ or a similar uncertainty measure.  If the intent is to estimate a well-calibrated infection distribution, then assuming a uniform $U(x)=1$ for all $x$ would be suitable.


\subsection{Better Utilization of Test-Kits}
In a country with a very large population (such as India having 1,376 million people), it might be extremely difficult to manufacture \textit{extra} test-kits in bulk. Thus, it is important to study how to optimize the use of these few extra test-kits. A recent study~\cite{test-kit-utilization} shows that it is possible to use lesser test-kits for testing more people in a situation where most of the test results turn out to be negative (number of positive cases in South Korea is only around $2.4\%$ of the total tests performed). The article~\cite{test-kit-utilization} mentions that researchers at the German Red Cross Blood Donor Service developed a procedure called \textit{mini-pool} where they mix $5$ trial samples in a pool and test it. If the test for the pool is negative, then all the individuals are declared negative. If the pool is tested positive, then each sample is tested individually. More sophisticated techniques can be used based on this binary search method.

Though this utilization problem is orthogonal to the problem we consider in this work, there are solutions that may help to choose the budget $K_i$, where $K_i$ can be much more than the number of available test-kits, and thus would help to test a larger set of individuals. 








\subsection*{Acknowledgements}
The authors thank Mohit Kumar and Arun Iyer for extremely useful discussions.

\bibliographystyle{plain}
\bibliography{main}

\begin{thebibliography}{10}

\bibitem{asha}
ASHA.
\newblock {Accredited Social Health Activist}.
\newblock
  \url{https://nhm.gov.in/index1.php?lang=1&level=1&sublinkid=150&lid=226}.

\bibitem{bbc2020coronavirus}
{BBC News}.
\newblock {Coronavirus: South Korea seeing a `stabilising trend'} {M}arch 15,
  2020.
\newblock
  \url{https://www.bbc.com/news/av/world-asia-51897979/coronavirus-south-korea-seeing-a-stabilising-trend?te=1&nl=the-interpreter&emc=edit\_int\_20200321}.

\bibitem{smoothing}
Cornell Course.
\newblock {CS 4740: Introduction to Natural Language Processing}.
\newblock
  \url{http://www.cs.cornell.edu/courses/cs4740/2014sp/lectures/smoothing+backoff.pdf}.

\bibitem{dudik2011doubly}
Miroslav Dud{\'\i}k, John Langford, and Lihong Li.
\newblock Doubly robust policy evaluation and learning.
\newblock {\em arXiv preprint arXiv:1103.4601}, 2011.
\newblock \url{https://arxiv.org/pdf/1103.4601.pdf}.

\bibitem{test-kit-utilization}
{EurekAlert}.
\newblock {Pool testing of SARS-CoV-02 samples increases worldwide test
  capacities many times over}.
\newblock
  \url{https://eurekalert.org/pub_releases/2020-03/guf-pto033020.php?fbclid=IwAR2G_XxXjp4hYEarCJxXBuQJfKSOLsk7prDop1imnQ0Ef0bFAYgMfRIuDiE}.

\bibitem{data-source}
Johns Hopkins University~Center for Systems~Science and Engineering~(JHU CSSE).
\newblock {2019 Novel Coronavirus COVID-19 (2019-nCoV) Data Repository}.
\newblock
  \url{https://raw.githubusercontent.com/CSSEGISandData/COVID-19/master/csse\_covid\_19\_data/csse\_covid\_19\_time_series/time\_series\_covid19\_confirmed\_global.csv}.

\bibitem{rate-limiting-google-cloud}
{Google Cloud}.
\newblock {Rate-limiting strategies and techniques}.
\newblock
  \url{https://cloud.google.com/solutions/rate-limiting-strategies-techniques}.

\bibitem{icmr2020}
{Indian Council of Medical Research (ICMR)}.
\newblock {Revised Strategy of COVID19 testing in India, Version 3, March 20,
  2020}.
\newblock
  \url{https://www.mohfw.gov.in/pdf/ICMRrevisedtestingstrategyforCOVID.pdf}.

\bibitem{krishnamurthy2019active}
Akshay Krishnamurthy, Alekh Agarwal, Tzu-Kuo Huang, Hal Daum{\'e}~III, and John
  Langford.
\newblock Active learning for cost-sensitive classification.
\newblock {\em Journal of Machine Learning Research}, 20(65):1--50, 2019.
\newblock
  \url{https://github.com/VowpalWabbit/vowpal\_wabbit/wiki/cs\_active.pdf}.

\bibitem{livemint2020icmr}
{LiveMint}.
\newblock {Introduction of community transmission of Covid-19 likely in India:
  ICMR study} {M}arch 24, 2020.
\newblock
  \url{https://www.livemint.com/news/india/introduction-of-community-transmission-of-covid-19-likely-in-india-icmr-study-11584992557087.html}.

\bibitem{nhp2019}
{National Health Profile}.
\newblock {Issue} 2019.
\newblock \url{https://www.cbhidghs.nic.in/showfile.php?lid=1147}.

\bibitem{nyt2020delay}
{The New York Times}.
\newblock {`It's Just Everywhere Already': How Delays in Testing Set Back the
  U.S. Coronavirus Response} {M}arch 10, 2020.
\newblock
  \url{https://www.nytimes.com/2020/03/10/us/coronavirus-testing-delays.html}.

\bibitem{nyt2020flatten}
{The New York Times}.
\newblock {The Coronavirus Outbreak.} {M}arch 23, 2020.
\newblock \url
  {https://www.nytimes.com/2020/03/23/world/asia/coronavirus-south-korea-flatten-curve.html}.

\bibitem{nyt2020asymptomatic}
{The New York Times}.
\newblock {They Were Infected With the Coronavirus. They Never Showed Signs}
  {F}ebruary 26, 2020.
\newblock
  \url{https://www.nytimes.com/2020/02/26/health/coronavirus-asymptomatic.html}.

\bibitem{selection-bias}
Arnau Tibau-Puig.
\newblock {Selection Bias: Origins and Mitigation}.
\newblock
  \url{https://www.quantifind.com/blog/selection-bias-origins-and-mitigation}.

\bibitem{tong2001support}
Simon Tong and Daphne Koller.
\newblock Support vector machine active learning with applications to text
  classification.
\newblock {\em Journal of machine learning research}, 2(Nov):45--66, 2001.

\bibitem{vowpal}
{Vowpal Wabbit}.
\newblock {Contextual Bandits Reinforcement Learning}.
\newblock \url{https://vowpalwabbit.org/tutorials/contextual\_bandits.html}.

\end{thebibliography}

\clearpage
\section{Appendix}


\begin{table}[!h] 
\caption{Sex ratios and persons above the age of 60, Statewise \cite{nhp2019} \label{StateAgeGender}}
\centering
\begin{tabular}{|c|r|r|r|r|}
\hline
\multirow{2}{*}{State} & \multirow{2}{*}{Sex Ratio} & \multicolumn{3}{c|}{Population above 60} \\ 		
&  & Total & Male & Female \\
\hline

India & 943 & 8.2 & 7.9 & 8.4\\ 
 Andhra Pradesh &-& 9.5 & 9.8 & 9.3\\ 
 Arunachal Pradesh & 938 &-&-&- \\ 
 Assam & 958 & 6.6 & 6.9 & 6.2\\ 
 Bihar & 918 & 6.4 & 6.4 & 6.3\\ 
 Chhattisgarh & 991 & 7.1 & 6.6 & 7.5\\ 
 Goa & 973 &  &  & \\ 
 Gujarat & 919 & 8.4 & 7.6 & 9.3\\ 
 Haryana & 879 & 7.4 & 6.9 & 8\\ 
 Himachal Pradesh & 972 & 11.1 & 10.8 & 11.4\\ 
 Jammu \& Kashmir & 889 & 9.3 & 9.2 & 9.3\\ 
 Jharkhand & 949 & 6.6 & 6.3 & 6.9\\ 
 Karnataka & 973 & 8.2 & 7.7 & 8.7\\ 
 Kerala & 1,084 & 13 & 12.3 & 13.7\\ 
 Madhya Pradesh & 931 & 7.2 & 6.9 & 7.5\\ 
 Maharashtra & 929 & 9.1 & 8.7 & 9.5\\ 
 Manipur & 985 &-&-&-\\ 
 Meghalaya & 989 &-&-&-\\ 
 Mizoram & 976 &-&-&-\\ 
 Nagaland & 931 &-&-&-\\ 
 Odisha & 979 & 9.9 & 10.1 & 9.8\\ 
 Punjab & 895 & 10.2 & 9.7 & 10.7\\ 
 Rajasthan & 928 & 7.3 & 6.5 & 8.2\\ 
 Sikkim & 890 &-&-&-\\ 
 Tamil Nadu & 996 & 10.7 & 10.6 & 10.9\\ 
 Telangana &-& 8.2 & 8.3 & 8.2\\ 
 Tripura & 960 &-&-&-\\ 
 Uttar Pradesh & 912 & 6.9 & 6.6 & 7.2\\ 
 Uttarakhand & 963 & 8.6 & 8 & 9.3\\ 
 West Bengal & 950 & 8.7 & 8.7 & 8.8\\ 
 Andaman \& Nicobar Islands & 876 &-&-&-\\ 
 Chandigarh & 818 &-&-&-\\ 
 Dadra \& Nagar Haveli & 774 &-&-&-\\ 
 Daman \& Diu & 618 &-&-&-\\ 
 NCT of Delhi & 868 & 6.6 & 6.4 & 6.9\\ 
 Lakshadweep & 947 &-&-&-\\ 
 Puducherry & 1,037 &-&-&-\\
\hline
\end{tabular}
\end{table}

\begin{table}[!h] 
\caption{National Age Distribution \cite{nhp2019} \label{AgeGender}}
\centering
\begin{tabular}{|c|c|c|c|}
\hline
Age Group & Total & Male & Female \\
\hline
 0-4  & 8.3  & 8.5  & 8.1\\ 
 5-9  & 8.8  & 8.9  & 8.6\\ 
 10-14  & 9.4  & 9.6  & 9.2\\ 
 15-19  & 10.2  & 10.5  & 10\\ 
 20-24  & 10.6  & 10.4  & 10.8\\ 
 25-29  & 9.8  & 9.7  & 10\\ 
 30-34  & 8.3  & 8.3  & 8.2\\ 
 35-39  & 7.2  & 7.1  & 7.2\\ 
 40-44  & 6.2  & 6.2  & 6.2\\ 
 45-49  & 5.3  & 5.3  & 5.3\\ 
 50-54  & 4.3  & 4.3  & 4.3\\ 
 55-59  & 3.5  & 3.4  & 3.6\\ 
 60-64  & 3  & 3  & 3.1\\ 
 65-69  & 2.1  & 2.1  & 2.2\\ 
 70-74  & 1.4  & 1.3  & 1.5\\ 
 75-79  & 0.9  & 0.8  & 0.9\\ 
 80-84  & 0.5  & 0.4  & 0.5\\ 
 85+  & 0.3  & 0.2  & 0.3 \\
\hline
\end{tabular}
\end{table}

\end{document}